\documentclass[10pt]{article}

\usepackage{times}
\usepackage{epsfig}
\usepackage{graphicx}
\usepackage{amsmath,amssymb}

\usepackage{algorithm}
\usepackage{listings}
\usepackage[table]{xcolor}
\usepackage{booktabs} %
\usepackage{colortbl}
\usepackage{glossaries}
\usepackage{multirow}
\usepackage[font=small]{caption}
\usepackage[a4paper, margin=1.5in]{geometry}
\usepackage{palatino}

\usepackage{pifont}%
\usepackage{xspace}
\usepackage{placeins}

\usepackage[pagebackref=true,breaklinks=true,letterpaper=true,colorlinks,bookmarks=false]{hyperref}

\newif\ifarxiv

\definecolor{myblue}{RGB}{8,48,107}
\definecolor{myred}{RGB}{127,39,4}

\newcommand{\cmark}{\ding{51}\xspace}%
\newcommand{\xmark}{\ding{55}\xspace}%
\newcommand{\xmarkg}{\textcolor{lightgray}{\ding{55}}\xspace}%

\newcommand{\mpm}[1]{\small(\textit{#1})}

\def \etal {\textit{et al.}\xspace}

\def \A   {\mathcal A}
\def \T   {\mathcal T}
\def \N   {\mathcal N}

\def \Bp {B}%
\def \Cp {C}%
\def \Dp {D}%
\def \ViTopt {T2}

\author{
\begin{minipage}{\linewidth}
\begin{center}
\large Ross Wightman$^{\circ}$ \hspace{0.25cm} Hugo Touvron$^{\star,\dagger}$ \hspace{0.25cm}Herv\'e J\'egou$^{\star}$ \\[0.5cm] 
\scalebox{0.88}{$^\circ$Independent researcher}\hspace{0.6cm} $^\star$Facebook AI\hspace{0.6cm} $^\dagger$Sorbonne University\\[1cm]
\end{center}
\end{minipage}
}
\title{ResNet strikes back: An improved training procedure in timm} 

\makeatletter
\let\inserttitle\@title
\makeatother

\makeatletter
\renewcommand{\paragraph}{%
  \@startsection{paragraph}{4}%
  {\z@}{2.8ex \@plus 1ex \@minus .2ex}{-1em}%
  {\normalfont\normalsize\bfseries}%
}
\makeatother

\date{~}
\begin{document}

\maketitle

\begin{abstract}
The influential Residual Networks designed by He et al. remain the gold-standard architecture in numerous scientific publications. They typically serve as the default architecture in studies, or as baselines when new architectures are proposed. 
Yet there has been significant progress on best practices for training neural networks since the inception of the ResNet architecture in 2015. Novel optimization \& data-augmentation have increased the effectiveness of the training recipes. 

In this paper, we re-evaluate the performance of the vanilla ResNet-50 when trained with a procedure that integrates such advances. 
We share competitive training settings and pre-trained models in the \textbf{timm} open-source library, with the hope that they will serve as better baselines for future work. For instance, with our more demanding training setting, a vanilla ResNet-50 reaches 80.4\% top-1 accuracy at resolution 224$\times$224 on ImageNet-val without extra data or distillation. 
We also report the performance achieved with popular models with our training procedure. 
\end{abstract}

\section{Introduction}
\label{sec:introduction} 

In the last decade we have witnessed significant advances in image classification, as reflected by improvement on benchmarks such as the ILSVRC'2012 challenge~\cite{Russakovsky2015ImageNet12} or other image classification benchmarks, which are visible on popular websites\footnote{See for instance \url{http://paperswithcode.com/task/image-classification}}.  
Schematically, the increase of performance reflects the maximization by the community of a problem of the form 
\begin{center}
accuracy (model) = $f (\A,\T,\N),$  
\end{center}
where $\A$ is the architecture design, $\T$ is the training setting along with its hyper-parameters, and $\N$ is the measurement noise, in which we also include overfitting that typically occurs when selecting the maximum over a large set of hyper-parameters or choices of methods. Several good practices exist to mitigate $\N$, like measuring the standard deviation with different seeds, using a separate evaluation dataset~\cite{Recht2019ImageNetv2} or evaluating models on transfer tasks. 
Putting aside $\N$, measuring progress on $\A$ or $\T$ poses a challenge as both $\A$ and $\T$ progress over time. When optimizing jointly over ($\A,\T$), there is no guarantee that the optimal choice $\T_1$ for a given architecture $\A_1$ remains the best for another model design $\A_2$. Therefore even when one compare models under the same training procedure, one may implicitly favor one model over another. One good practice to disentangle the improvement resulting from the training procedure from that of the architecture is to ensure that the baseline incorporates new ``ingredients'' from the literature, and to put a reasonable amount of effort in adjusting the hyper-parameters. Ideally, i.e., without resource and time constraints, one would optimally adopt the best possible training procedure for each architecture 
\begin{equation}
    \T^\star(\A) = \max_{\T} f(\A,\T,\N), %
    \label{equ:model_optim}
\end{equation}
but realistically this is not possible. When comparing architectures, most papers compare their results to other reported in older publications, but for which architectures were trained with potentially weaker recipes. %
In the best case, the same or a similar procedure is used to compare two architectures. %
 
We are not aware of an effort specifically targeted at improving the ResNet-50 training procedure with an extensive ingredient selection and hyper-parameter search. In the literature, the performance reported on ImageNet-1k-val for this architecture ranges from 75.2\% to 79.5\%, depending on the paper. It is unclear whether a sufficient effort has been invested in pushing the baseline further.  
We want to fill this gap: 
in this paper, we focus on the vanilla ResNet-50 architecture\footnote{ResNet-50 V1.5 (the PyTorch~\cite{pytorch} ResNet50) a slight adjustment of He \etal~\cite{He2016ResNet} that was made in \href{http://torch.ch/blog/2016/02/04/resnets.html}{torch7}, stride was moved from 1x1 to 3x3 in bottleneck. } as described by He \etal~\cite{He2016ResNet}, and we optimize the training so as to maximize the performance of this model for the original test resolution of $224\times224$.
We solely consider the training recipe. Therefore we exclude all variations of the ResNet-50 such as SE-ResNet-50~\cite{Hu2017SENet} or ResNet-50-D~\cite{he2019bag}, which usually improve the accuracy under the same training procedure. 
In summary, in this paper, 
\begin{itemize}
    
    \item We propose three training procedures intended to be strong baselines for a vanilla ResNet-50 used at inference resolution $224\times 224$. The three variants correspond to different numbers of epochs (100, 300 and 600) with adjustment of hyper-parameters and ingredients.  
    
    \item Our procedure include recent advances from the literature as well as new proposals. Noticeably, we depart from the usual cross-entropy loss. Instead, our training solves a multi-classification problem when using Mixup and CutMix: we minimize the binary cross entropy for each concept selected by these augmentations, assuming that all the mixed concepts are present in the synthetized image. 
    
    \item We measure the stability of the accuracy over a large number of runs with different seeds, and discuss the overfitting issue by jointly comparing the performance on ImageNet-val versus the one obtained in ImageNet-V2~\cite{Recht2019ImageNetv2}. 

    \item We train popular architectures and re-evaluate their performance. We also discuss the necessity to  optimize jointly the architecture and the training procedure: we showcase that having the same training procedure is not sufficient for comparing the merits of different architectures.  

\end{itemize}

We provide ablations in Section~\ref{sec:ablations}.  Our supplemental material may interest the community: 
Appendix~\ref{sec:augreg} details augmentations variants that have been introduced by the \textbf{timm} library\footnote{available at \url{http://github.com/rwightman/pytorch-image-models/}}. 
Appendix~\ref{sec:alternative} covers alternative procedures for training a ResNet-50 that significantly differ in their ingredients from our three focal training procedures. They achieve noteworthy performance and possibly better results with different architectures and tasks.

\section{Related work}
\label{sec:related} 

\paragraph{Image Classification} is a core problem in computer vision. It is often employed as a benchmark task %
to measure progress in computer vision. Pre-trained models for image classification, particularly trained on ImageNet~\cite{deng2009imagenet}, are used in a large variety of downstream tasks like detection or segmentation. 
Progress in image classification generally translates to progress on these tasks. 

\paragraph{The timm library~\cite{pytorchmodels}} has recently gained significant momentum in the scientific community as it provides implementations for numerous popular models for image classification, as well as training methods. Pre-trained weights -- either adapted from originals or trained in timm with newer procedures -- are included for many models. While model architectures are \textbf{timm}’s focus, it also includes implementations of many data augmentations, regularization techniques, optimizers, and learning rate schedulers that are leveraged in the training procedures described in this paper. In many cases these implementations include functionality beyond the original implementations or papers that they were based upon. We describe these additions in Appendix~\ref{sec:augreg}. %

\paragraph{ResNet~\cite{He2016ResNet}} is one of the most popular image classification architectures. It was a noteworthy improvement at the time it was introduced and continues to serve as the referent architecture for some analysis ~\cite{Cubuk2019RandAugmentPA,Yun2019CutMix,Zhang2017Mixup}, or as a baseline in papers introducing new architectures~\cite{Radosavovic2020RegNet,Ridnik2020TResNetHP,xie2017aggregated,zhang2020resnest}. 

Some works have modernized the ResNet training procedure and obtained some improvement over the original model (e.g. Dollar \etal~\cite{Dollr2021FastAA}). 
This allows a more direct comparison when considering new models or methods involving more elaborate training procedures than the one initially used.
Nevertheless, improving the ResNet-50 baseline~\cite{caron2021emerging,Dollr2021FastAA,he2019bag,Touvron2020GrafitLF,Touvron2019FixRes,Yuan2021TokenstoTokenVT} was not the main objective of these works. 
As a consequence and as we will see, the best performance reported so far with a ResNet-50 is still far from the maximum performance (peak or average) that one can achieve with this architecture.
In this paper, our goal is to offer the best possible training procedure that we could find for the ResNet-50 based on existing ingredients and practices. We hope that it will serve as a strong baseline for subsequent works. 
Note, some papers have also focused on ResNet-50 training~\cite{Bello2021RevisitingRI,Lee2020CompoundingTP,vaswani2021scaling}, but they have either modified the architecture or changed the resolution, which does not allow for a direct comparison to the original ResNet-50 at resolution 224$\times$224. For instance, Lee~\etal~\cite{Lee2020CompoundingTP} use ResNet-D~\cite{he2019bag} with SE attention~\cite{Hu2017SENet}. 
Bello \etal \cite{Bello2021RevisitingRI} also optimize ResNet without architectural changes, but they don't report competitive results for ResNet-50 at  224$\times$224.

\paragraph{Training ingredients \& recipes}
for image classification have significantly evolved since the inception of AlexNet~\cite{Krizhevsky2012AlexNet}. 
Several trends have changed over time. 
Common modifications include replacing the waterfall schedule (classical division by 10 of learning rate every 30 epochs) by a longer and more progressive schedule~\cite{Brock2021HighPerformanceLI,Dollr2021FastAA,dosovitskiy2020image,tan2019efficientnet,Touvron2020TrainingDI,Yalniz2019BillionscaleSL}. 
Increasing jointly the number of epochs~\cite{Dollr2021FastAA,dosovitskiy2020image,he2019bag,tan2019efficientnet,Touvron2020TrainingDI} and the batch size while using mixed precision better leverages powerful GPUs.  
Modern procedures make use of stronger data-augmentation~\cite{Ekin2018AutoAugment,Cubuk2019RandAugmentPA,Yun2019CutMix,Zhang2017Mixup,Zhong2020RandomED}, stronger regularization~\cite{Foret2021SharpnessAwareMF,Huang2016DeepNW,srivastava2014dropout}, 
weight averaging~\cite{Izmailov2018AveragingWL,Polyak1992AccelerationOS} and correct the train-test resolution discrepancy~\cite{Touvron2019FixRes} by differentiating the train from the test resolution~\cite{Brock2021HighPerformanceLI,dosovitskiy2020image}. %
Different losses have also been experimented with~\cite{Beyer2020ImageNetReal,Khosla2020SupervisedCL} even if cross-entropy remains the standard.
For the optimization, SGD with Nesterov momentum~\cite{Sutskever2013OnTI} is a common default for CNNs. RMSProp is also used for specific CNN architecture families like in Inception~\cite{Szegedy2016RethinkingTI}, NASNet~\cite{Zoph2018LearningTA}, AmoebaNet~\cite{Real2018AmobeaA}, MobileNet~\cite{Howard2017MobileNetsEC}, EfficientNet~\cite{tan2019efficientnet} .%
For training image classifiers based on transformers~\cite{dosovitskiy2020image,Touvron2020TrainingDI} and MLP~\cite{tolstikhin2021MLPMixer,Touvron2021ResMLPFN}, AdamW~\cite{Loshchilov2017AdamW} and Lamb~\cite{you20lamb} optimizers are popular choices.

\section{Training Procedures}
\label{sec:method} 

We offer three different training procedures with different costs and performance so as to cover different use-cases, see Table~\ref{tab:cost} for resource usage and corresponding accuracies. 
Our procedures target the best performance of ResNet-50 when tested at resolution $224\times224$. We have explored numerous variations with different optimizers, choice of regularization, and a reasonable amount of grid search for the hyper-parameters. 
We refer the reader to Section \ref{sec:overfitting} for control experiments on quantifying the amount of overfitting. 
See Section~\ref{sec:ingredient_details} in the Appendix for the exact ingredient list and parametrization. 
We focus on three different operating points:
\begin{description}
    \item [Procedure A1] aims at providing the best performance for ResNet-50. It is therefore the longest in terms of epochs (600) and training time (4.6 days on one node with 4 V100 32GB GPUs). 
    \item [Procedure A2] is a 300 epochs schedule that is comparable to several modern procedures like DeiT, except with a larger batch size of 2048 and other choices introduced for all our recipes. 
    \item [Procedure A3] aims at outperforming the original ResNet-50 procedure with a short schedule of 100 epochs and a batch size 2048. It can be trained in 15h on 4 V100 16GB GPUs and could be a good setting for exploratory research or studies. 
\end{description}

Note, Section~\ref{sec:alternative} gives alternative training procedures that may serve as interesting choices when considering other models. In the rest of this section, we focus on the ingredients included in A1--A3.

\begin{table}[t]
\begin{center}
    \scalebox{0.87}{
    \begin{tabular}{c|rcrcc|ccc}
    \toprule
         Training  & Number & Training   & Training  & Peak memory  & Numbers &\multicolumn{3}{c}{Top-1 accuracy} \\
         Procedure & of epochs & resolution & time \ \  & by GPU (MB)  & of GPU & val & real & v2 \\
         \midrule
         A1 & 600 \ \    & $224\times224$ & 110h\ \ &    22,095    & 4            & 80.4 & 85.7 & 68.7 \\
         A2 & 300 \ \    &  $224\times224$&  55h\ \ &  22,095    & 4            & 79.8 & 85.4 & 67.9\\
         A3 & 100 \ \    & $160\times160$ &  15h\ \ &   11,390    & 4            & 78.1 & 84.5 & 66.1\\
    \bottomrule
    \end{tabular}}
\end{center}
    \caption{Training resources used for our three training procedures on V100 GPUs and corresponding accuracies at resolution 224$\times$224 on ImageNet1k-val, -V2 and -Real. 
    Note, the top-1 val acc. of pytorch-zoo~\cite{pytorch} is 76.1\%. 
    \label{tab:cost}}
    \vspace{-15pt}
\end{table}

\paragraph{Loss: multi-label classification objective.} 
Mixup and CutMix augmentation synthesize an image from several images having in most cases different labels. By using cross-entropy, the output is implicitly treated as a probability of presence of each of the mixed concepts. In our training, we assume instead that these concepts are all present, and treat the classification as a multi-label classification problem (1-vs-all). 
For this purpose, we adopt the binary cross-entropy (BCE) loss instead of the typical cross-entropy (CE).
This loss is consistent with the Mixup and CutMix data augmentation: The targets are defined for each class to 1 (or $1-\varepsilon$ with smoothing) if the class is selected by Mixup or Cutmix, independent of other classes.
Over the best settings that we have explored, BCE slightly outperforms cross-entropy in their best respective configurations. 
We point out that Beyer et al.~\cite{Beyer2020ImageNetReal} previously adopted BCE with the motivation to produce multiple non-exclusive labels, and obtained excellent results with it. But to the best of our knowledge they did not use it with CutMix or Mixup as we propose to do. 
In our experiments, even when using BCE, setting all mixed concepts with a target to 1 (or $1-\varepsilon$) is more effective than considering a distribution of concepts that sum to 1. Conceptually we believe it is more aligned with what Mixup and CutMix are actually doing: it is likely that a human could recognize each of two mixed concepts.

\paragraph{Data-Augmentation.}
We adopt the following combination of data augmentations: 
on top of standard Random Resized Crop (RRC) and horizontal flip (commonly used since GoogleNet ~\cite{Szegedy2015Goingdeeperwithconvolutions}), we apply \textbf{timm}~\cite{pytorchmodels} variants of RandAugment~\cite{Cubuk2019RandAugmentPA}, Mixup~\cite{Zhang2017Mixup}, and CutMix~\cite{Yun2019CutMix}. This combination was used for instance in DeiT~\cite{Touvron2020TrainingDI}. Many of the model weights in \textbf{timm} have also been trained with RandAugment and Mixup, but with Random Erasing~\cite{Zhong2020RandomED} and increased regularization instead of CutMix. We refer the reader to Appendix~\ref{sec:augreg} for more details about the variants offered in \textbf{timm}.

\paragraph{Regularization.} \quad
Across our three training procedures, regularization differs the most. In addition to adapting the weight decay, we use label smoothing, Repeated-Augmentation~\cite{berman2019multigrain,hoffer2020augment} (RA) and stochastic-Depth~\cite{Huang2016DeepNW}. We use more regularization for longer training schedules. For instance we adopt label smoothing only for A1. Both RA and stochastic depth tend to improve the results at convergence, but they slow down the training in the early stages as reported by Berman et al.~\cite{berman2019multigrain} for RA. For short schedules they are therefore less effective or even detrimental, which is why we adopt them only with A1 and A2. %
Note that for other architectures, or larger ResNets, %
it is beneficial to add additional regularization, therefore one would have to adapt the corresponding hyper-parameters for such architectures. %
For instance, for a ResNet-152 the performance increases from 81.8\% to 82.4\% on Imagenet-val by putting more of RandAugment, mixup and stochastic depth regularization on top of A2 recipe. At resolution 256$\times$256 this model obtains 82.7\%, which is above the accuracy (82.2\%)   reported by Bello \etal~\cite{Bello2021RevisitingRI} for a ResNet-200 before architectural changes (Table 1 in their paper).

\paragraph{Optimization.} \quad
Since AlexNet, the most used optimizer to train convnets is SGD. 
In contrast transformers~\cite{dosovitskiy2020image,touvron2021going} and MLP~\cite{tolstikhin2021MLPMixer,Touvron2021ResMLPFN} use AdamW~\cite{Loshchilov2017AdamW} or LAMB optimizer. 
Dosovitskiy et al.~\cite{dosovitskiy2020image} report similar performance between AdamW~\cite{Loshchilov2017AdamW} and SGD for ResNet-50.
This concurs with our observations for intermediate batch sizes (e.g., 512). 
We use larger batches, e.g., 2048. When combined with repeated augmentation and the binary cross entropy loss, we found that LAMB~\cite{you20lamb} makes it easier to consistently achieve good results. We found it difficult to achieve convergence when using both SGD and BCE.  
We therefore focus on LAMB with cosine schedule as the default optimizer for training our ResNet-50. Alternative training procedures using different optimizer, loss, augmentation, and regularization combinations can be found in Appendix~\ref{sec:alternative}. %

\begin{table}
\caption{
 Ingredients and hyper-parameters used for ResNet-50 training in different papers. We compare existing training procedures with ours. %
\label{tab:comp_hyperparameters}}
\centering
\scalebox{0.78}
{%
\begin{tabular}{l|ccccc|ccc}
\toprule
 & \multicolumn{5}{c|}{Previous approaches} & \multicolumn{3}{c}{Ours} \\
\cmidrule{2-9}
Procedure $\rightarrow$ & 
ResNet& 
PyTorch &
FixRes & 
DeiT & 
FAMS ($\times$4) & 
A1& A2 & A3 \\
Reference & 
\cite{He2016ResNet}& 
\cite{pytorch} &
\cite{Touvron2019FixRes} & 

\cite{Touvron2020TrainingDI} & 
\cite{Dollr2021FastAA}
\\
\midrule
Train Res &
224 & 
224 &
224 & 

224 & 
224 & 
224   & 224  & 160\\
Test Res  &
224 &
224 &
224 &

224 &
224 & 
224   & 224  & 224 \\
\midrule
Epochs   &
90 & 
90 &
120 &

300 & 
400 & 
600   & 300  & 100\\
\# of forward pass &
450k &
450k & 
300k &

375k & 
500k &
375k & 
188k & 
63k\\ 
\midrule
Batch size & 
256 & 
256 &
512 & 

1024 & 
1024  & 
2048 & 2048 & 2048\\
Optimizer &
SGD-M & 
SGD-M & 
SGD-M &

AdamW & 
SGD-M &
LAMB & LAMB & LAMB \\
LR      & 
0.1 & 
0.1 &
0.2 & 

$1\times 10^{-3}$  & 
2.0 & 
$5\times 10^{-3}$ &  $5\times 10^{-3}$ &  $8\times 10^{-3}$ \\
LR decay& 
step  &
step & 
step & 

cosine &  
step  & 
cosine & cosine & cosine  \\
 decay rate & 
 0.1 & 
 0.1 &
 0.1  &

 \_ & 
 $0.02^{t/400}$ &
 \_ & \_ & \_ \\
 decay epochs & 
 30 & 
 30 & 
 30 & 

 \_
 & 1 
 & \_ & \_ & \_ \\
Weight decay     &
$10^{-4}$  & 
$10^{-4}$  & 
$10^{-4}$ &  

0.05 & 
$10^{-4}$ &
0.01 &
0.02 &
0.02    \\
Warmup epochs & 
\xmarkg  & 
\xmarkg & 
\xmarkg & 

5 & 
5  &
5   &  5 & 5   \\
\midrule
Label smoothing $\varepsilon$ & 
\xmarkg & 
\xmarkg &
\xmarkg & 

0.1  &
0.1  & 
0.1 & \xmarkg & \xmarkg  \\
Dropout      & 
\xmarkg & 
\xmarkg & 
\xmarkg & 

\xmarkg  & 
\xmarkg & 
\xmarkg  &  \xmarkg &  \xmarkg\\
Stoch. Depth & 
\xmarkg & 
\xmarkg & 
\xmarkg &

0.1 & 
\xmarkg & 
0.05 &
0.05 &
\xmarkg\\
Repeated Aug & 
\xmarkg & 
\xmarkg & 
\cmark &

\cmark &
\xmarkg & 
\cmark & \cmark & \xmarkg\\
Gradient Clip. & 
\xmarkg  & 
\xmarkg & 
\xmarkg & 

\xmarkg & 
\xmarkg & 
\xmarkg & \xmarkg & \xmarkg \\
\midrule
H. flip  & 
\cmark & 
\cmark &
\cmark & 

\cmark & 
\cmark  &
\cmark & \cmark & \cmark\\
RRC & 
\xmarkg & 
\cmark & 
\cmark & 

\cmark &  
\cmark & 
\cmark & \cmark & \cmark \\
Rand Augment  &
\xmarkg & 
\xmarkg &
\xmarkg &

9/0.5 &
\xmarkg& 
7/0.5 & 7/0.5 & 6/0.5\\
Auto Augment  & 
\xmarkg & 
\xmarkg & 
\xmarkg & 

\xmarkg & 
\cmark & 
\xmarkg & \xmarkg & \xmarkg  \\

Mixup alpha  & 
\xmarkg & 
\xmarkg & 
\xmarkg & 

0.8 &
0.2 & 
0.2  & 0.1 & 0.1  \\
Cutmix alpha &
\xmarkg  & 
\xmarkg  & 
\xmarkg  & 

1.0    &
\xmarkg & 
1.0   & 
1.0  & 
1.0 \\
Erasing prob. &
\xmarkg   &
\xmarkg   &
\xmarkg   & 

0.25 &
\xmarkg  &
\xmarkg  & \xmarkg   & \xmarkg \\
ColorJitter  & 
\xmarkg & 
\cmark & 
\cmark & 
\xmarkg & 

\xmarkg &
\xmarkg & \xmarkg & \xmarkg  \\
PCA lighting  & 
\cmark &
\xmarkg &
\xmarkg & 

\xmarkg & 
\xmarkg & 
\xmarkg & \xmarkg & \xmarkg \\
\midrule
SWA & 
\xmarkg & 
\xmarkg &
\xmarkg &

\xmarkg & 
\cmark & 
\xmarkg & \xmarkg & \xmarkg  \\
EMA &
\xmarkg & 
\xmarkg & 
\xmarkg & 

\xmarkg & 
\xmarkg &
\xmarkg &  \xmarkg & \xmarkg \\
\midrule
Test  crop ratio & 
0.875 & 
0.875 &
0.875 & 

0.875 &
0.875  &
0.95 &
0.95 &
0.95\\
\midrule
CE loss  & 
\cmark &  
\cmark & 
\cmark & 

\cmark &
\cmark & 
\xmarkg & 
\xmarkg & 
\xmarkg\\
BCE loss &
\xmarkg & 
\xmarkg & 
\xmarkg &

\xmarkg & 
\xmarkg & 
\cmark & 
\cmark & 
\cmark  \\
\midrule
Mixed precision & 
\xmarkg &
\xmarkg &
\xmarkg & 

\cmark  & 
\cmark & 
\cmark & 
\cmark & 
\cmark \\
\midrule
Top-1 acc. &
75.3\% &
76.1\% &
77.0\% &

78.4\% & 
79.5\% & 
80.4\% & 79.8\% & 78.1\% \\

 \bottomrule
\end{tabular}}
\end{table}

\paragraph{Details of our ingredients and comparison to existing training procedures. }
In Table~\ref{tab:comp_hyperparameters} we compare different recipes used to train vanilla ResNet-50 to ours. 
We consider only the results with the unmodified ResNet-50 architecture. 
We have chosen a wide range of training procedures to try to be as representative as possible but obviously it cannot be exhaustive. 
We do not consider approaches using advanced 
training settings like distillation, or models pre-trained self-supervised or with pseudo-labels.

\section{Experiments}
\label{sec:experiments} 

In this section we first compare our training procedure to existing ones and evaluate them with different architectures. 
Importantly, we discuss the significance of our results with experiments that aim at (1) quantifying the sensitivity of the performance to random factors; (2) evaluating the overfitting by measuring on a different test set. 

\subsection{Comparison of training procedures for ResNet-50} %
\label{sec:ingredient_details}

Table~\ref{tab:cost} summarizes the main characteristics of our training procedure. To the best of our knowledge, our procedure A1 surpasses the current state of the art on ImageNet with a vanilla ResNet-50 architecture at resolution 224$\times$224. Our other procedures A2 and A3 achieve lower but still high performance with less resources.  

\paragraph{Performance comparison with other architectures.}

In Table~\ref{tab:mainres} we report the performance obtained when training
different architectures with our training procedures. This allows us to see how well they generalize to other models.
Or procedures improves the performance of several models over results reported in the literature, in particular older ones and/or those most comparable to ResNet-50 in terms of architecture and size. In some cases like ViT-B, we observe that A2 is better than A1, which suggests that the hyper-parameters are not adapted to longer schedules (typically requiring more regularization). For instance, the A2 training recipe achieves 81.8\% top-1 accuracy when training a ResNet-152, but by increasing a bit the regularization we improved it to 82.4\% at resolution 224$\times$224, which translates to 82.7\% when evaluated at resolution 256$\times$256.%

In Table~\ref{tab:mainres}, we compare the performance and resources associated with our 3 training recipes when using them to train other architectures. 
We complement these results with Table~\ref{tab:efficiency_models}, where we additionally include the performance and efficiency on ImageNet-1k, ImageNet-V2 and ImageNet-Real for different architectures trained with our best performing A1 training recipes.

\begin{table}[t]
\caption{
Comparison on ImageNet classification between other architectures trained with our ResNet-50 optimized training procedure \textbf{without any hyper-parameters adaptation}. In particular, our procedure must be adapted for deeper/larger models, which benefit from more regularization.
For the training cost we report the training time (time) in hours, the number of GPU used (\#GPU) and the peak memory by GPU (Pmem) in GB.
For A1 and A2, we adopt the same training and test resolution as in the original publication introducing the architecture. For A3 we use a smaller training resolution to reduce the compute-time. 
$^\dagger$: torchvision~\cite{pytorch} results. $^*$: DeiT~\cite{Touvron2020TrainingDI} results. 
\label{tab:mainres}
}
    \scalebox{0.62}{
    \begin{tabular}{l@{\, }|rr|cc|rr|rr|rrr|ccc|c}
        \toprule
                        &  \multicolumn{2}{c}{A1-A2-org.}  &  \multicolumn{2}{c}{A3}& \multicolumn{7}{|c}{Cost} & \multicolumn{4}{|c}{ImageNet-1k-val}\\
                \cmidrule{2-16}
                                & train & test                     & train & test& A1 & A2 & \multicolumn{2}{|c}{A1-A2} & \multicolumn{3}{|c|}{A3}  & A1 & A2 & A3 & org. \\
                                \cmidrule{6-16}
           $\downarrow$  Architecture       & res. & res.                      & res. & res.& \multicolumn{2}{c|}{time (hour)}  & \# GPU & Pmem & time & \# GPU & Pmem  & \multicolumn{4}{c}{Accuracy(\%)} \\
        \midrule
        ResNet-18~\cite{He2016ResNet}$^\dagger$ & 224 & 224 & 160 & 224 & 186 & 93 & 2 & 12.5  & 28 & 2 & 6.5 & 71.5 & 70.6 & 68.2 & 69.8 \\
        ResNet-34~\cite{He2016ResNet}$^\dagger$ & 224 & 224 & 160 & 224 & 186 & 93 & 2 & 17.5  & 27 & 2 & 9.0 & 76.4 & 75.5 & 73.0 & 73.3 \\
        \rowcolor{blue!10}
        ResNet-50~\cite{He2016ResNet}$^\dagger$ & 224 & 224 & 160 & 224 & 110 & 55 & 4 & 22.0 & 15 & 4 & 11.4 & 80.4 & 79.8 & 78.1 & 76.1 \\
        ResNet-101~\cite{He2016ResNet}$^\dagger$ & 224 & 224 & 160 & 224 & 74 & 37 & 8 & 16.3 & 8 & 8 & 8.5 & 81.5 & 81.3 & 79.8 & 77.4 \\
        ResNet-152~\cite{He2016ResNet}$^\dagger$ & 224 & 224 & 160 & 224 & 92  & 46 & 8 & 22.5 & 9 & 8 & 11.8 & 82.0 & 81.8 & 80.6 & 78.3\\
        \midrule
        RegNetY-4GF~\cite{Radosavovic2020RegNet} & 224 & 224 & 160 & 224 & 130 & 65 & 4 & 27.1 & 15 & 4 & 13.9 & 81.5 & 81.3 & 79.0 & 79.4\\
        RegNetY-8GF~\cite{Radosavovic2020RegNet} & 224 & 224 & 160 & 224 & 106  & 53 & 8 & 19.8 & 10 & 8 & 10.3 & 82.2 & 82.1 & 81.1 & 79.9\\
        RegNetY-16GF~\cite{Radosavovic2020RegNet} & 224 & 224 & 160 & 224 & 150  & 75 & 8 & 25.6 & 13 & 8 & 13.4 & 82.0 & 82.2 & 81.7 & 80.4\\
        RegNetY-32GF~\cite{Radosavovic2020RegNet} & 224 & 224 & 160 & 224 & 120  & 60 & 16 & 17.6 & 12 & 16 & 9.4 & 82.5 & 82.4 & 82.6 & 81.0\\
        \midrule
        SE-ResNet-50~\cite{Hu2017SENet}& 224 & 224 & 160 & 224 & 102  & 51 & 4 & 27.6 & 16 & 4 & 14.2 & 80.0 & 80.1 & 77.0 & 76.7 \\
        SENet-154~\cite{Hu2017SENet}& 224 & 224 & 160 & 224  & 110 & 55 & 16 & 23.3 & 12 & 16 & 12.2 & 81.7 & 81.8 & 81.9 & 81.3\\

        \midrule
        ResNet-50-D~\cite{he2019bag}& 224 & 224 & 160 & 224 & 100  & 50 & 4 & 23.9 & 14 & 4 & 12.3 & 80.7 & 80.2 & 78.7 & 79.3 \\
        ResNeXt-50-32x4d~\cite{xie2017aggregated}$^\dagger$& 224 & 224 & 160 & 224  & 80 & 40 & 8 & 14.3 & 15 & 4 & 14.6 & 80.5 & 80.4 & 79.2 & 77.6\\
        \midrule
        EfficientNet-B0~\cite{tan2019efficientnet} & 224 & 224 & 160 & 224 & 110  & 55 & 4 & 22.1 & 15 & 4 & 11.4 & 77.0 & 76.8 & 73.0 & 77.1\\
        EfficientNet-B1~\cite{tan2019efficientnet} & 240 & 240 & 160 & 224 & 62  & 31  & 8 & 17.9 & 8 & 8 & 7.9 & 79.2 & 79.4 & 74.9 & 79.1\\ 
        EfficientNet-B2~\cite{tan2019efficientnet} & 260 & 260 & 192 & 256 & 76  & 38 & 8 & 22.8 & 9 & 8 & 11.9 & 80.4 & 80.1 & 77.5 & 80.1 \\ 
        EfficientNet-B3~\cite{tan2019efficientnet} & 300 & 300 & 224 & 288 & 62  & 31 & 16 & 19.5 & 6 & 16 & 10.1 & 81.4 & 81.4 & 79.2 & 81.6\\ 
        EfficientNet-B4~\cite{tan2019efficientnet} & 380 & 380 & 320 & 380 & 64  & 32 & 32 & 20.4 & 8 & 32 & 14.3 & 81.6 & 82.4 & 81.2 & 82.9\\ 

        \midrule
        ViT-Ti~\cite{Touvron2020TrainingDI}$^*$& 224 & 224 & 160 & 224  & 98 & 49 & 4 & 16.3 &  14 & 4 & 7.0 & 74.7 & 74.1 & 66.7 & 72.2 \\
        ViT-S~\cite{Touvron2020TrainingDI}$^*$ & 224 & 224 & 160 & 224  & 68 & 34 & 8& 16.1 & 8 & 8  & 7.0 & 80.6 & 79.6 & 73.8 & 79.8\\
        ViT-B~\cite{dosovitskiy2020image}$^*$  & 224 & 224 & 160 & 224  & 66 & 33 & 16 & 16.4 &  5 & 16 & 7.3 & 80.4 & 79.8 & 76.0 & 81.8\\
        \toprule
         \multicolumn{16}{c}{\textbf{timm}~\cite{pytorchmodels} specific architectures} \\
        \midrule
        ECA-ResNet-50-T & 224 & 224 & 160 & 224 & 112  & 56 & 4 & 29.3 & 15 & 4 & 15.0 & 81.3 & 80.9 & 79.6 & \_\\
        \midrule
        EfficientNetV2-rw-S~\cite{Tan2021EfficientNetV2SM} & 288 & 384 & 224 & 288 & 52  & 26 & 16 & 16.6 & 7 & 16 & 10.1 & 82.3 & 82.9 & 80.9 & 83.8\\ 
        EfficientNetV2-rw-M~\cite{Tan2021EfficientNetV2SM} & 320 & 384 & 256 & 352 & 64  & 32 & 32 & 18.5 & 9 & 32 & 12.1 & 80.6 & 81.9 & 82.3 & 84.8\\ 
        \midrule
        ECA-ResNet-269-D & 320 & 416 & 256 & 320 & 108  & 54 & 32 & 27.4 & 11 & 32 & 17.8 & 83.3 & 83.9 & 83.3 & 85.0\\ 
        \bottomrule
    \end{tabular}}
\end{table}

\begin{table}[t]

    \caption{
\textbf{Performance of models trained with A1 training procedure.} We measure peak memory and  throughput on one GPU V100 32GB with batch size 128, FP16 precision and test resolution from Table~\ref{tab:mainres}.
Note that the throughput is indicative, since it depends on the GPU hardware, the software that runs the models, and other factors like the adjustment of batch size (we keep it fix in this table). 
\label{tab:efficiency_models}
}
    \centering
    \smallskip
    \scalebox{0.85}{
    \begin{tabular}{|l|rr|rr|ccc|}
        \toprule
                      & \# params    & FLOPs       & Throughput  & Peak mem & Top-1 & Real & V2 \\
         Architecture& $\times 10^6$ & $\times 10^9$ & (im/s)    & (MB)       & Acc. & Acc. & Acc.\\
        \midrule
        ResNet-18~\cite{He2016ResNet} & 11.7 & 1.8 & 7960.5 & 588 & 71.5 & 79.4 & 59.4\\
        ResNet-34~\cite{He2016ResNet} & 21.8 & 3.7 & 4862.6 & 642 & 76.4 & 83.4 & 65.1\\
        \rowcolor{blue!10}
        ResNet-50~\cite{He2016ResNet} & 25.6 & 4.1 & 2536.6 & 1,155 & 80.4 & 85.7 & 68.7\\
        ResNet-101~\cite{He2016ResNet} & 44.5 & 7.9 & 1547.9 & 1,264 & 81.5 & 86.3 & 70.3\\
        ResNet-152~\cite{He2016ResNet} & 60.2 & 11.6 & 1094.0 & 1,355 & 82.0 & 86.4 & 70.6\\
        \midrule
        RegNetY-4GF~\cite{Radosavovic2020RegNet} & 20.6 & 4.0 & 1690.6 & 1,585 &81.5 & 86.7 & 70.7\\
        RegNetY-8GF~\cite{Radosavovic2020RegNet} & 39.2 & 8.1 & 1122.3 & 2,139 & 82.2 & 86.7 & 71.1\\
        RegNetY-16GF~\cite{Radosavovic2020RegNet} & 83.6 & 16.0 & 694.1 & 3,052 & 82.0 & 86.4 & 71.2\\
        RegNetY-32GF~\cite{Radosavovic2020RegNet} & 145.0 & 32.4 & 431.5 & 3,366 & 82.5 & 86.6 & 71.7\\
        \midrule
        SE-ResNet-50~\cite{Hu2017SENet} & 28.1 & 4.1 & 2174.8 & 1,193 & 80.0 & 85.8 & 68.8\\
        SENet-154~\cite{Hu2017SENet} & 115.1 & 20.9 & 511.5 & 2,414 & 81.7 & 86.0 & 71.2\\
        \midrule
        ResNet-50-D~\cite{he2019bag} & 25.6 & 4.4 & 2418.8 & 1,205 & 80.7 & 85.9 & 68.9\\
        ResNeXt-50-32x4d~\cite{xie2017aggregated} & 25.0 & 4.3 & 1727.5 & 1,247 & 80.5 & 85.5 & 68.4\\

        \midrule
        EfficientNet-B0~\cite{tan2019efficientnet} & 5.3 & 0.4 & 3701.5 & 932 & 77.0 & 83.8 & 65.0\\
        EfficientNet-B1~\cite{tan2019efficientnet} & 7.8 & 0.7 & 2365.2 & 1,077 & 79.2 & 85.3 & 67.7\\
        EfficientNet-B2~\cite{tan2019efficientnet} & 9.2 & 1.0 & 1786.8 & 1,318 & 80.4 & 86.0 & 69.3\\
        EfficientNet-B3~\cite{tan2019efficientnet} & 12.0 & 1.8 & 1082.4 & 2,447 & 81.4 & 86.7 & 70.4\\
        EfficientNet-B4~\cite{tan2019efficientnet} & 19.0 & 4.2 & 561.3 & 5,058 & 81.6 & 85.9 & 70.8\\

        \midrule
        ViT-Ti~\cite{Touvron2020TrainingDI}& 5.7 & 1.3 & 3497.7 & 346 & 74.7 & 82.1 & 62.4\\
        ViT-S~\cite{Touvron2020TrainingDI} & 22.0 & 4.6 & 1762.3 & 682 & 80.6 & 85.6 & 69.4\\
        ViT-B~\cite{dosovitskiy2020image}  & 86.6 & 17.6 & 771.0 & 1,544 & 80.4 & 84.8 & 69.4\\

        \midrule
        \multicolumn{8}{|c|}{\textbf{timm}~\cite{pytorchmodels} specific architectures} \\
        \midrule
        ECA-ResNet50-T & 25.6 & 4.4 & 2139.7 & 1,155 & 81.3 & 86.1 & 69.9\\
        \midrule
        EfficientNetV2-rw-S~\cite{Tan2021EfficientNetV2SM} & 23.9 & 8.8 & 823.1 & 2,339 & 80.6 & 84.8 & 69.2\\
        EfficientNetV2-rw-M~\cite{Tan2021EfficientNetV2SM} & 53.2 & 18.5 & 456.8 & 2,916 & 82.3 & 87.1 & 71.7\\
       \midrule
        ECA-Resnet269-D & 102.1 & 70.6 & 168.1 & 4,134 & 83.3 & 86.9 & 71.9\\
        \bottomrule
    \end{tabular}}
\end{table}

\subsection{Significance of measurements: seed experiments}  
\label{sec:overfitting}
\begin{figure}[t]
              \includegraphics[width = 0.43\linewidth]{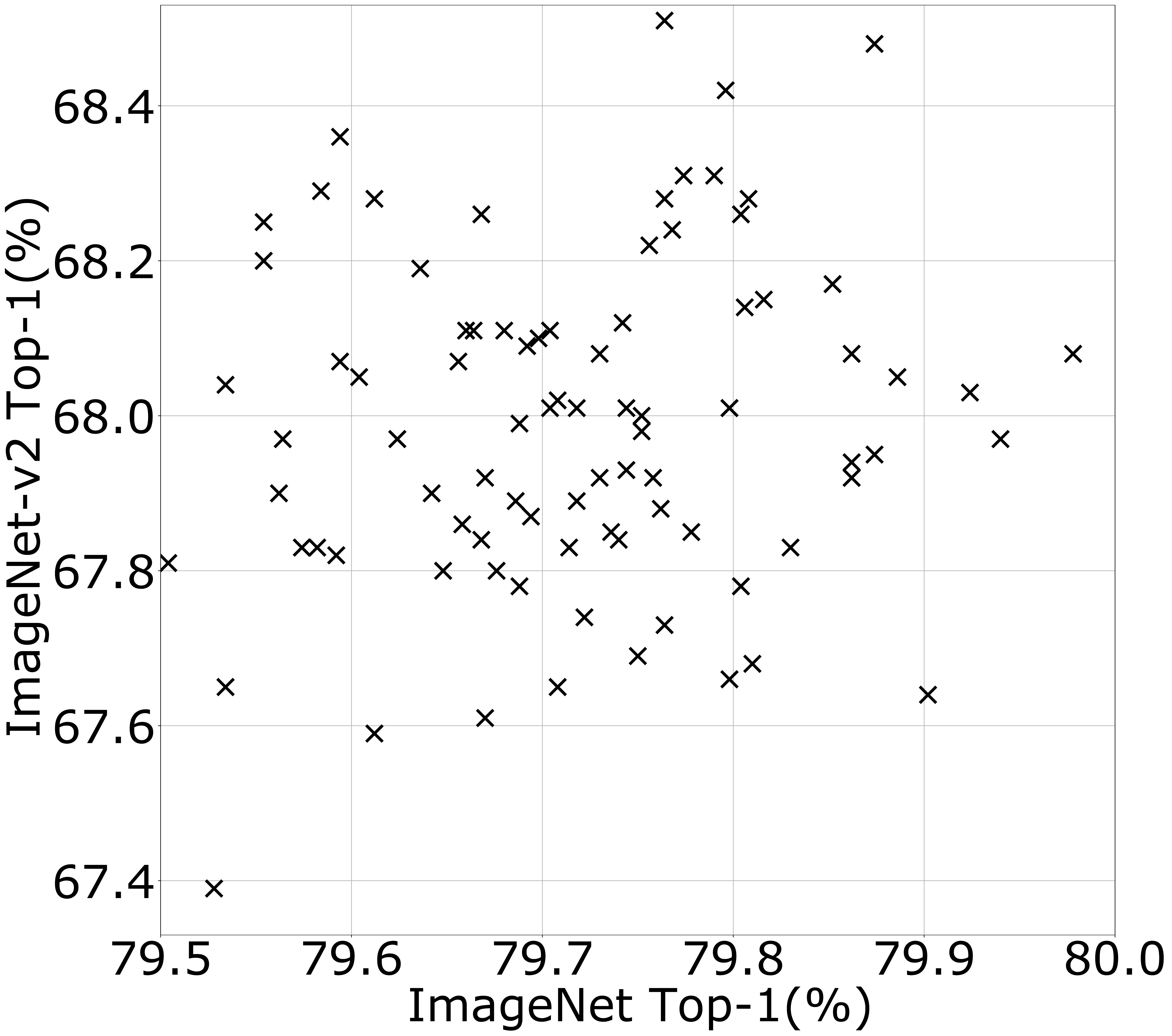}
             \hfill
    \raisebox{2.5cm}{
    \begin{minipage}{0.54\linewidth}
    \scalebox{0.8}{
    \begin{tabular}{l|cc|cc|c}
         \toprule
          & \multicolumn{5}{c}{Top-1 accuracy (\%)}  \\
         \cmidrule{2-6}
         dataset $\downarrow$ & mean & std & max & min & seed 0 \\
         \midrule
         ImageNet-val  & 79.72 & 0.10 & 79.98 & 79.50 & 79.85 \\
         ImageNet-real & 85.37 & 0.08 & 85.55 & 85.21 & 85.45 \\
         ImageNet-V2   & 67.99 & 0.23 & 68.69 & 67.39 & 67.90 \\
      \bottomrule
    \end{tabular}}
    \caption{\emph{Top $\uparrow$:} Statistics for  ResNet-50 trained with A2 and 100 different seeds.
    The column "seed 0" corresponds to the weights that we take as reference. Its performance is +0.13\% above the average top-1 accuracy on Imagenet-val. 
    \smallskip \newline
    \emph{$\leftarrow$ Left:} Point cloud plotting the ImageNet-val top-1  accuracy vs ImageNet-V2 for all seeds. Note that the outlying seed that achieves 68.5\% top-1 accuracy on ImageNet-V2 has an average performance on ImageNet-val. 
    \label{fig:seed_xps1}
    }
   \end{minipage}}
\end{figure}

For a fixed set of choices and hyper-parameters, there is some inherent variability on the performance due to the presence of random factors in several stages. 
It is the case for the weight initialization, but also for the optimization procedure itself. For instance the order in which the images are fed to the network through batches depends on a random generator. This variability raises the question of the significance of accuracy measurements. 
For this purpose, we measure the distribution of performance when changing the random generator choices. This is conveniently done by changing the seed, as previously done by Picard~\cite{picard21luckyseed}, who concludes to the exist of outliers significantly outperforming or underperforming the average outcome of a traing procedure. 
In Figure~\ref{fig:seed_xps1}, %
we report several statistics on the performance with the A2 training procedure when considering 100 distinct seeds (from 1 to 100, note that we have used seed=0 in all other experiments).
In these experiments, we focus on the performance reached at the end of the training: we do not select the maximum obtained by intermediate checkpoints in the last epochs. This would have a similar effect as a seed selection, but the measures would not be IID and less disentangled from the training duration itself.

The standard deviation is typically around $0.1$ on ImageNet-val, see Figure~\ref{fig:seed_xps1}. This concurs with statistics reported in the literature for ResNet and other convnets~\cite{Radosavovic2020RegNet}. 
The variance is higher on ImageNet-V2 (std=0.23), which consists of a smaller set (10000 vs 50000 for -val) of  images not present in the validation set. 
The mean 79.72\% shows that our main weights (seed 0) overestimates the average performance by about $+0.13\%$.

\begin{figure}[h]
             \includegraphics[width = 0.41\linewidth]{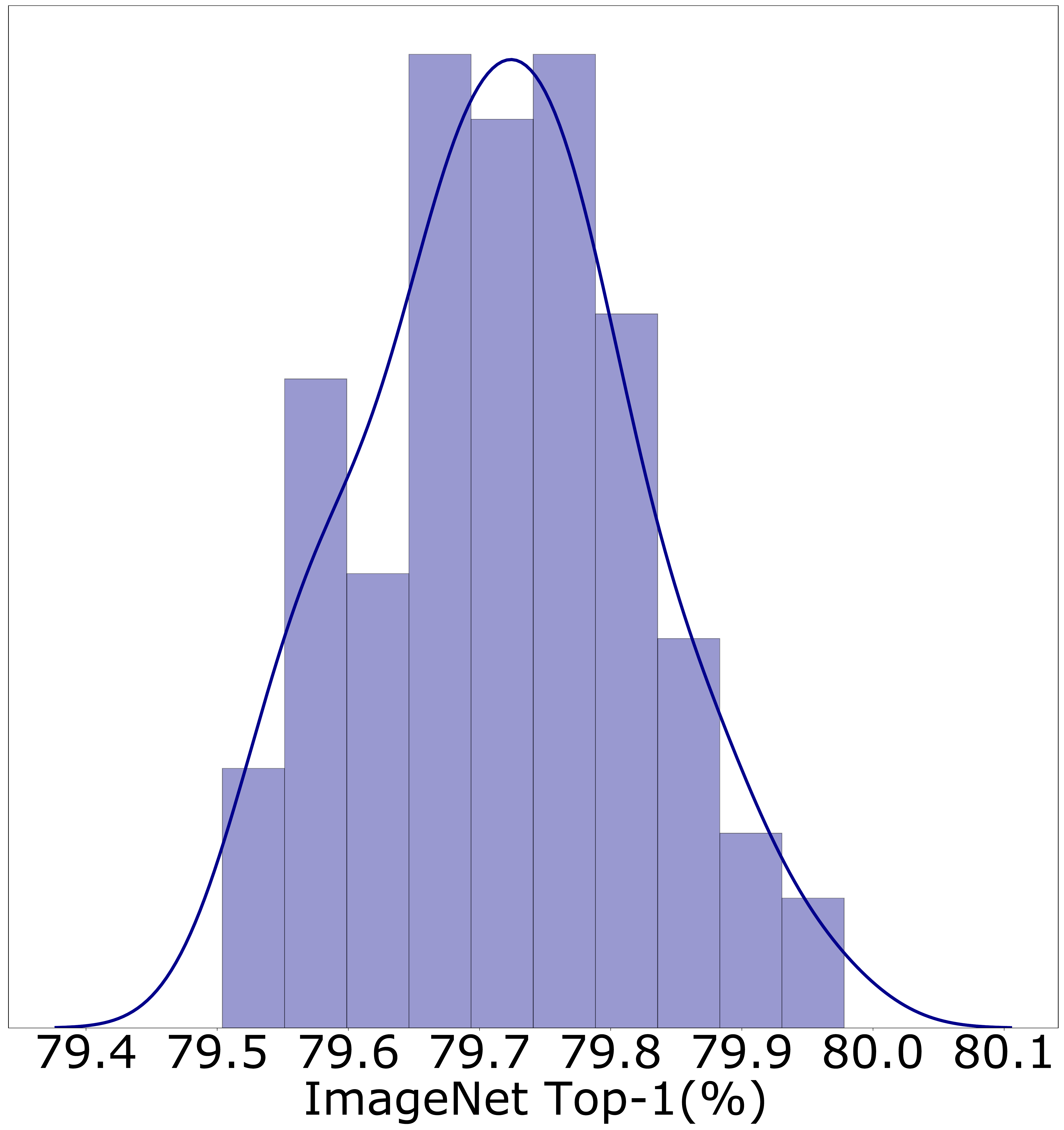}
             \hfill
    \raisebox{2.95cm}{
    \begin{minipage}{0.54\linewidth}
    \caption{Distribution of the performance on ImageNet-val with the A2 procedure. It is measured with 100 different seeds. We also depict the Gaussian-fit of this distribution. 
    \label{fig:seed_histo}
    }
   \end{minipage}}
\end{figure}

\paragraph{Peak performance and control of overfitting} To prevent to over-estimate too much the accuracy on validation, during our exploration process we have selected only the final checkpoint and we use relatively coarse grid for hyper-parameters search to prevent introducing an additional seed effect. 
However optimizing over a large number of choices typically leads to overfitting. In Figure~\ref{fig:seed_xps1}, we observe that the maximum (or peak performance) is close to 80.0\% with the A2 training procedure.  Note, Figure~\ref{fig:seed_histo} provides the distribution of accuracy as an histogram;

One question is whether this model is intrinsically better than the average ones, or if it was just lucky on this particular measurement set. To attempt to answer this question, we measure how the performance transfers to another measurement dataset: we compute for all the seeds the couples (ImageNet-val top-1 acc., ImageNet-V2 top-1 acc.), and plot them as a point cloud in Figure~\ref{fig:seed_xps1}. We observe that the correlation between the performance on ImageNet-val and -V2 is limited. Noticeably the best performance is not achieved by the same seed on the two datasets.  
This observation  suggests some significant measurement noise, which advocates to report systematically the performance on different datasets, and more particularly one making a clear distinction between validation and test.

\paragraph{More on sensitivity analysis: variance along epochs.}

\begin{figure}[h]
        \includegraphics[width = 0.48\linewidth]{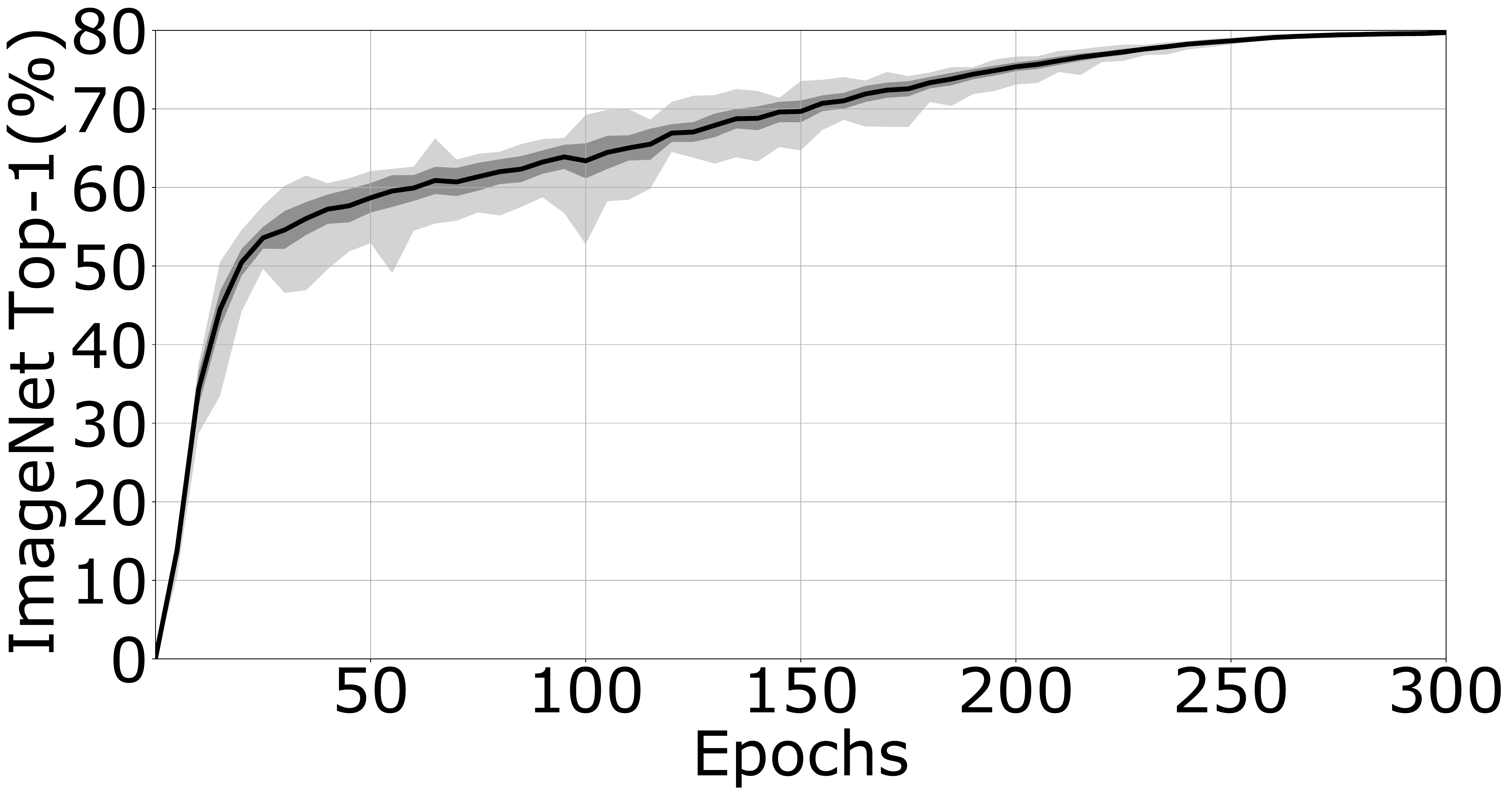} \hfill
        \includegraphics[width = 0.48\linewidth]{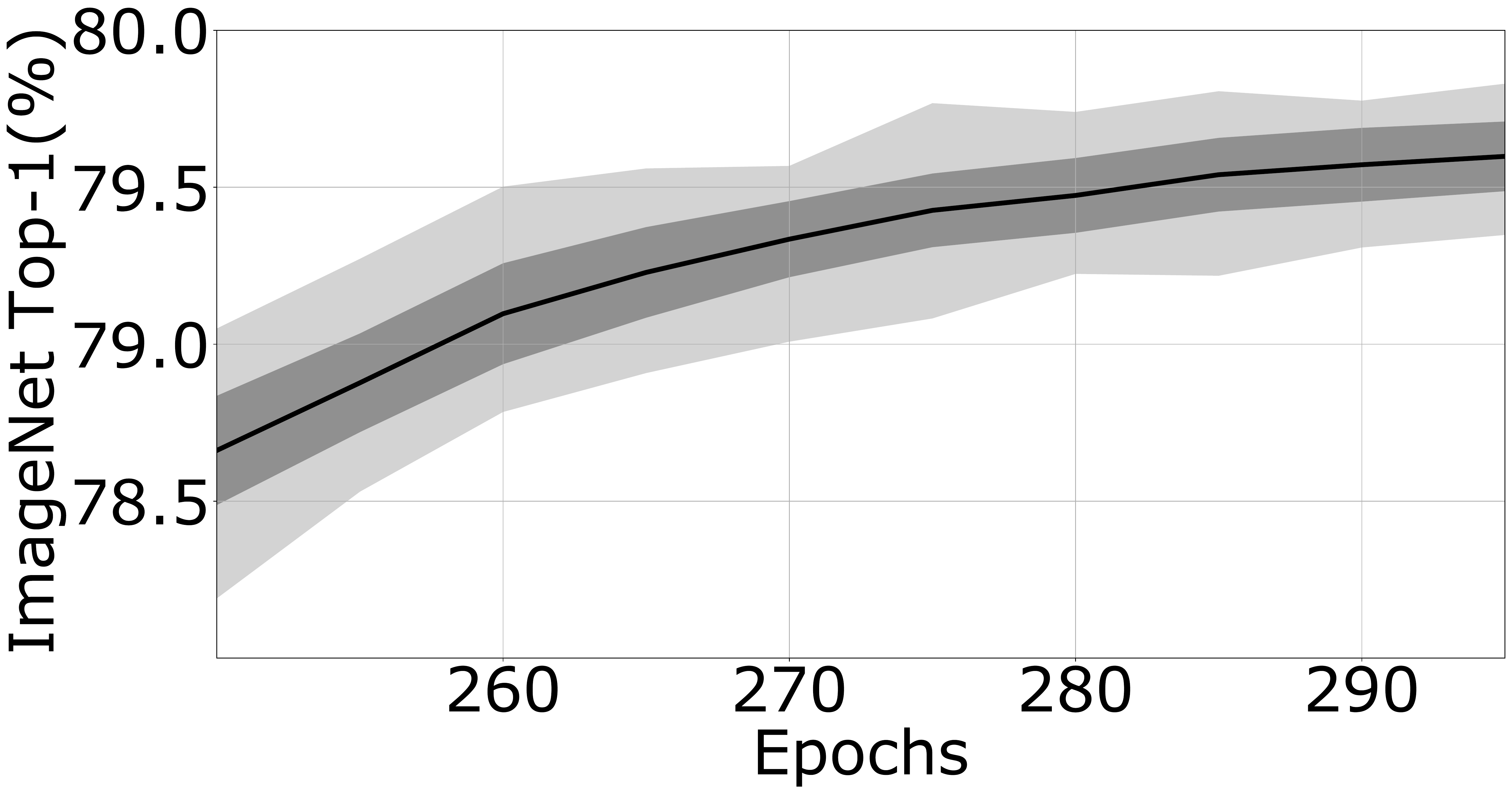}\\
    \caption{We show how the mean, standard deviation, minimum and maximum of the top-1 accuracy on ImageNet-val evolves during training with the A2 procedure (ResNet-50 architecture). \textbf{(Left)} For all 300 training epochs. \textbf{(Right)} Same but for the last epochs. 
    We note that the variance in accuracy is high at the beginning, see for instance at epoch 100, where the difference in performance can be as large as 10\% in accuracy. Towards the end of the training, most of the networks converge to similar values and the range significantly decreases in the last 50 epochs. 
    \emph{Credit}: this figure and experiment was inspired by Picard~\cite{picard21luckyseed}. 
    \label{fig:training_curve}}
\end{figure}

Figure~\ref{fig:training_curve} shows how the performance variability evolves along epochs. 

\subsection{Transfer Learning}

In Table~\ref{tab:transfer} we provide  transfer learning performance on seven fine grained dataset with our different pre-training procedures, and provide a comparison  with the default PyTorch pre-training. 
For each pre-training we use exactly the same fine-tuning procedure inspired by the fine-tuning procedure used in DeiT~\cite{Touvron2020TrainingDI}.
For each dataset we adapt the fine-tuning hyper-parameters.

We observe that the fine-tuning tend to smooth the difference of performance on certain datasets, such as CIFAR or Stanford Cars. Overall our A1 procedure leads to the best performance on downstream tasks, but the performance of the Pytorch default and A2 tend to be similar, while on Imagenet-val and -v2 A2 was significantly better. A3 is significantly inferior on downstream tasks, which may be related to the lower training resolution at 160$\times$160. 

\begin{table}
\caption{Performance comparison on transfer-learning tasks for different pre-training recipes.  \label{tab:transfer}}
\centering
{\small
\begin{tabular}{lrrr|c|ccc|}
\toprule
Dataset & Train size & Test size & \#classes & Pytorch~\cite{pytorch} & A1 & A2 & A3  \\
\midrule
ImageNet-val \cite{Russakovsky2015ImageNet12}  & 1,281,167 & 50,000 & 1000 & 76.1 & \textbf{80.4} & 79.8 & 78.1 \\ 
iNaturalist 2019~\cite{Horn2019INaturalist}& 265,240   & 3,003  & 1,010 & 73.2 & 73.9 & \textbf{75.0} & 73.8 \\ 
Flowers-102~\cite{Nilsback08}& 2,040 & 6,149 & 102 & \textbf{97.9} & \textbf{97.9} & \textbf{97.9} & 97.5 \\ 
Stanford Cars~\cite{Cars2013}& 8,144 & 8,041 & 196 & 92.5 & \textbf{92.7} & 92.6 & 92.5 \\  
CIFAR-100~\cite{Krizhevsky2009LearningML}  & 50,000    & 10,000 & 100 & 86.6 & \textbf{86.9} & 86.2 & 85.3   \\ 
CIFAR-10~\cite{Krizhevsky2009LearningML}  & 50,000    & 10,000 & 10 & 98.2 & \textbf{98.3} & 98.0 & 97.6  \\ 
\bottomrule
\end{tabular}}
\end{table}

\subsection{Comparing architectures and training procedures: a show-case of contradictory conclusions} 

In this paragraph we case how difficult it is to compare two architectures, even under the same training procedure, or conversely how it is difficult to compare different procedures with a single architecture. 
We choose ResNet-50 and DeiT-S. The latter~\cite{Touvron2020TrainingDI} is essentially a ViT parameterized so that it has approximately the same number of parameters as a ResNet-50. For each architecture, we have put a significant effort in optimizing the procedure to maximize the performance on Imagenet-val with the same 300 epochs training schedule and same batch size. Under this constraint, the best training procedure that we have designed for ResNet-50 is A2. We denote by \ViTopt~ the corresponding training procedure for DeiT-S. Note that this training procedure achieves a significantly better performance on Imagenet-val than the one initially proposed for DeiT-S (80.4\% versus 79.8\% in the original paper). 

\def \Te {\mathcal T} 
\begin{center}
{\small
\begin{tabular}{lrrrr}
\toprule

  \hspace{2.6cm} test set $\rightarrow$      &  \multicolumn{2}{c}{ImageNet-val}  &  \multicolumn{2}{c}{ImageNet-v2} 
  \\[4pt]
  \cmidrule(l){2-3}   \cmidrule(l){4-5} 
$\downarrow$\, architecture \ \ \ training $\rightarrow$ & A2\ \ & \ViTopt\ \  & A2\ \  & \ViTopt\ \   \\
\midrule
ResNet-50   & \cellcolor{blue!10} 79.9  &  79.2      & \cellcolor{blue!10} 67.9   & 67.9 \\    
DeiT-S      &  79.6                     & \cellcolor{blue!10} 80.4  & 68.1   & \cellcolor{blue!10} 69.2 \\
\bottomrule
\end{tabular}}
\end{center}

As one can see, by choosing the procedure optimized for any of the two architectures, one may conclude that this architecture is better based on ImageNet-val accuracy: with A2 training, ResNet50 is better than DeiT-S, with T2 training, DeiT-S is better than ResNet50. The measurements on ImageNet-v2 would lead to a different conclusion, as DeiT-S is better for both procedure. But even in that case, by focusing on A2 one may conclude that the difference between ResNet-50 and DeiT-S with A2 training is not statistically significant: 67.9\% vs 68.1\%. 
Conversely, if the goal is to compare A2 to \ViTopt, we could draw different conclusions on ImageNet-val if considering a single architecture. 

\section{Ablations}
\label{sec:ablations}

In this section we provide a few ablations of hyper-parameters or selection of ingredients.  
Some modifications are difficult to ablate individually since they require to re-adjust several other parameters to work properly.   
This is the case in particular of the optimizer, which strongly interacts with other choices and hyper-parameters.  
In Appendix~\ref{sec:alternative} we provide alternative training procedures that we have developed for other optimizers: RMSProp, SGD and AdamP. %

\begin{table}[t]
\begin{center}
\scalebox{0.78}{
\begin{tabular}{cccc|ccc} %
\toprule
loss     &  LR & WD & RA  & A2   \\
\midrule
\quad BCE & $2\times10^{-3}$ & $0.02$ & \cmark  & 78.24    \\
\quad BCE & $2\times10^{-3}$ & $0.03$ & \cmark  & 78.47    \\
\quad BCE & $3\times10^{-3}$ & $0.02$ & \cmark  & 79.16    \\
\quad BCE & $3\times10^{-3}$ & $0.03$ & \cmark  & 79.28    \\
\quad BCE & $5\times10^{-3}$ & $0.01$ & \cmark  & 79.66     \\
\rowcolor{blue!10}
\quad BCE & $5\times10^{-3}$ & $0.02$ & \cmark  & 79.85     \\
\quad BCE & $5\times10^{-3}$ & $0.03$ & \cmark  & 79.73    \\
\quad BCE & $8\times10^{-3}$ & $0.02$ & \cmark  & 79.63     \\
\midrule
\quad BCE & $3\times10^{-3}$ & $0.02$ & \xmark  & 78.74    \\
\quad BCE & $5\times10^{-3}$ & $0.02$ & \xmark  & 79.57    \\
\quad BCE & $5\times10^{-3}$ & $0.03$ & \xmark  & 79.58    \\
\midrule
\quad CE & $2\times10^{-3}$  & $0.02$ & \cmark  & 77.37    \\
\quad CE & $3\times10^{-3}$  & $0.02$ & \cmark  & 78.22    \\
\quad CE & $5\times10^{-3}$  & $0.02$ & \cmark  & 79.18    \\
\quad CE & $5\times10^{-3}$  & $0.03$ & \cmark  & 79.23    \\
\quad CE & $5\times10^{-3}$  & $0.05$ & \cmark  & 79.31    \\
\quad CE & $8\times10^{-3}$  & $0.03$ & \cmark  & 79.12    \\

\midrule
\quad CE & $3\times10^{-3}$  & $0.02$ & \xmark  & 77.71   \\
\quad CE & $5\times10^{-3}$  & $0.01$ & \xmark  & 78.93   \\
\quad CE & $5\times10^{-3}$  & $0.02$ & \xmark  & 79.00   \\
\quad CE & $5\times10^{-3}$  & $0.03$ & \xmark  & 78.62    \\
\quad CE & $8\times10^{-3}$  & $0.02$ & \xmark  & 78.72   \\
\bottomrule
\end{tabular}}
\end{center}
\smallskip 
\caption{Main ablation table with A2 procedure. We compare BCE vs CE, including repeated augmentation or not, and vary the learning rate LR and weight decay WD in ranges that our exploration phase has identified as being the most adapted. All results are reported with Seed 0 and therefore all the ResNet-50 are initialized with the same weights when the training starts.  \colorbox{blue!10}{The highlighted row} corresponds to our A2 procedure. 
\label{tab:main_ablation}}
\end{table}

\paragraph{Main ingredients and hyper-parameters.} 
In Table~\ref{tab:main_ablation} we provide an ablation of major ingredients. We focus on the intermediate A2 training procedure as it is a good compromise between compute-cost  and accuracy. 
We make the following observations:
\begin{itemize}
    \item \emph{Learning rate and Weight Decay.} The learning rate has an important effect on performance. The higher value $5.10^{-3}$ presented in this table leads to the best performance. However increasing it further increases the risk of divergence. 
    We have typically set the weight decay in the range [0.02, 0.03] that we have identified in our preliminary exploration. 
    This parameter is a bit sensitive and can interact with other forms of regularization. In some cases we observe significant differences between 0.02 and 0.03. 
    \item \emph{Loss: Binary Cross Entropy versus Cross Entropy.} In this ablation, moving back from how we use BCE to the vanilla CE loss significantly reduces the performance. 
    As discussed in our main paper, we use the flexibility of BCE to regard Mixup/Cutmix as activating a multi-class 1-vs-all classification problem as discussed in our paper, as opposed to the choice of enforcing probabilities that sum to 1. If we enforce probabilities to sum to 1 as implicitly done with cross-entropy, we obtain a slightly lower accuracy as reported in Table~\ref{tab:ablation_augmentation}. By itself, i.e., with the same target, we do not conclude that BCE is necessarily better than CE. But it is with that loss that we reach the configuration with the highest accuracy overall. 

    \item \emph{Repeated augmentation } is providing a small boost in this ablation. This augmentation has some complex interaction with other hyper-parameters, and is not well understood in our opinion. In some cases we observed that it was neutral or detrimental, for instance with shortest schedules (A3 procedure), or in Table~\ref{tab:ablation_augmentation} with higher values of the Mixup parameter. Overall, it was best to include this ingredient in our most accurate procedures A1 and A2. 
\end{itemize}

\begin{table}[t]
\centering
\scalebox{0.78}{
\begin{tabular}{l|rrr}
\toprule
drop-factor   & A1 \quad & A2 \quad &  A3 \quad \\
\midrule
\quad 0       & 79.94  &  79.79  & \cellcolor{blue!10} 78.06    \\
\quad 0.05    & \cellcolor{blue!10} 80.38 & \cellcolor{blue!10} 79.85  & 77.57    \\
\quad 0.1     & 80.12  & 79.62  & 77.32    \\
\bottomrule
smoothing     & \\
\ \ \quad \xmark  & 80.22 & \cellcolor{blue!10} 79.85  & \cellcolor{blue!10} 78.06\\
\ \ \quad \cmark  & \cellcolor{blue!10} 80.38 & 79.58 & 77.99\\
\bottomrule
\end{tabular}}
\smallskip 
\caption{Ablation of stochastic Depth \& smoothing for our training procedures. In \colorbox{blue!10}{blue}, we highlight the results corresponding to the default selection for each procedure, see Table~\ref{tab:comp_hyperparameters}.\label{tab:ablation_stochastic_depth}}
\end{table}

\paragraph{Stochastic Depth \& Smoothing.}  

We have included stochastic depth in the A1 and A2 training procedures. In Table~\ref{tab:ablation_stochastic_depth} we observe that it provides an improvement for A2 compared to setting the drop-rate to 0 (i.e., no stochastic depth), not for A3. 
Label smoothing is not effective at 300 epochs with other hyper-parameters and ingredients fixed. This is why we only use it for the longer 600-epoch schedule in A1, where our exploration concluded that it has a positive impact. 

\paragraph{Augmentation.} 
Table~\ref{tab:ablation_augmentation} evidences the role of augmentations when we modify a few parameters (of Mixup and RandAugment): each modification that we have done has some impact on the measured score. 
While it would be unrealistic (and not ecological) to ensure that all our choices are statistically significant, one can observe that all modifications in this table   decrease the top-1 accuracy below the average performance (79.72\% -- std {\small$0.1$}) that we report over 100 seeds in Figure~\ref{fig:seed_xps1}.

\begin{table}[h]
\centering
\scalebox{0.85}{
\begin{tabular}{lccclc|c}
\toprule
mixup	& Rep. aug.& RandA	 & label smooth.	& stoch. depth & BCE target & top-1 acc.\\
\midrule
\rowcolor{blue!10} 
\quad 0.1     & \cmark  & 7 & \xmark & \quad 0.05  & \cmark & 79.85  \\
\quad 0.2     & \xmark  &   &        & \quad       &        & 79.62  \\
\quad 0.2     &         & 6 &        & \quad       &        & 79.61  \\
\quad 0.05    &         &   &        & \quad       &        & 79.57  \\
\quad         &         &   &        & \quad       & \xmark & 79.57  \\

\bottomrule
\end{tabular}}
\smallskip 
\caption{Ablation of some data-augmentation choices for our training procedure A2 on Imagenet-val, all computed with "Seed 0". 
The first row contains our default choices, see Table~\ref{tab:comp_hyperparameters} for the full set of hyper-parameters. Each other row corresponds to an ablation for which we have changed only one or two hyper-parameters or ingredient. Activating "BCE target" is our default. It refers to our choice to regard Mixup/Cutmix as activating a multi-class classification 1-vs-all problem as discussed in our paper. Not using it means that we also use BCE, but we enforce the probabilities of the concepts sum to 1 as with the regular cross-entropy loss. 
\label{tab:ablation_augmentation}}
\end{table}

\paragraph{Crop-ratio.} We evaluate the influence of the crop-ratio used at inference time. The one most commonly adopted in the literature is 0.875. Recently researchers have considered larger values for this parameter, noticeably for vision transformers after significant gains were reported by the author of the \textbf{timm} library with these models. 
Table~\ref{tab:ablation_crop_ratio} provides an analysis as a function of this parameter.

\begin{table}[h]
\begin{center}
\scalebox{0.73}{
\begin{tabular}{l|c@{\ \ \ }c@{\ \ \ }r|c@{\ \ \ }c@{\ \ }r|c@{\ \ \ }c@{\ \ \ }r}
\toprule
    & \multicolumn{3}{c}{A1}              & \multicolumn{3}{c}{A2}   & \multicolumn{3}{c}{A3}\\
    \cmidrule(lr){2-4} \cmidrule(lr){5-7} \cmidrule(lr){8-10}
    crop-ratio & mean \mpm{std}  & max\,--\,min  & seed 0  & mean   \mpm{std}  & max\,--\,min      & seed 0  & mean   \mpm{std}  & max\,--\,min      & seed 0  \\
    \cmidrule(lr){1-1}     \cmidrule(lr){2-4} \cmidrule(lr){5-7} \cmidrule(lr){8-10}
\quad 0.875      & 80.18  \mpm{0.14} & 80.45\,--\,79.90  & 80.14 & 79.67 \mpm{0.08} & 79.91\,--\,79.59  & 79.91 & 77.69 \mpm{0.10} & 77.85\,--\,77.48 & 77.69\\
\quad 0.9        & 80.22  \mpm{0.15} & \textbf{80.54}\,--\,79.98  & 80.25 & 79.73 \mpm{0.09} & 79.89\,--\,79.56  & 79.75 & 77.86  \mpm{0.09}       & 78.01\,--\,77.62 & 77.83       \\
\quad 0.95       & 80.24  \mpm{0.14} & 80.49\,--\,79.91  & \cellcolor{blue!10} 80.38 & 79.68 \mpm{0.09} & 79.85\,--\,79.57  & \cellcolor{blue!10} 79.85 & 78.00 \mpm{0.09} & 78.09\,--\,77.83 & \cellcolor{blue!10} 78.06       \\
\quad 1.0        & 80.15  \mpm{0.11} & 80.15\,--\,79.66  & 80.19 & 79.58 \mpm{0.13} & 79.88\,--\,79.32  & 79.88                     & 78.02 \mpm{0.10}  & 78.16\,--\,77.83 & 77.93      \\
\bottomrule
\end{tabular}}
\end{center}
\caption{Ablation of the crop-ratio when training with A1. We compute the Imagenet-val top-1 accuracy as a function of this parameter for 10 different seeds, for ResNet50 trained with our procedures. 
Our selection of 0.95 was based on Seed 0 in early experiments. It is comparable but not statistically better than the standard 0.875. 
Note that we have one A1 seed that leads to a top 80.54\% top-1 accuracy at crop-ratio 0.9. We regard it as being overfit and therefore we do not recommend to report this number. 
\label{tab:ablation_crop_ratio}}
\end{table}

\paragraph{Evaluation at other resolutions.}

\begin{figure}[t]
    \centering
    \includegraphics[width = 0.5\linewidth]{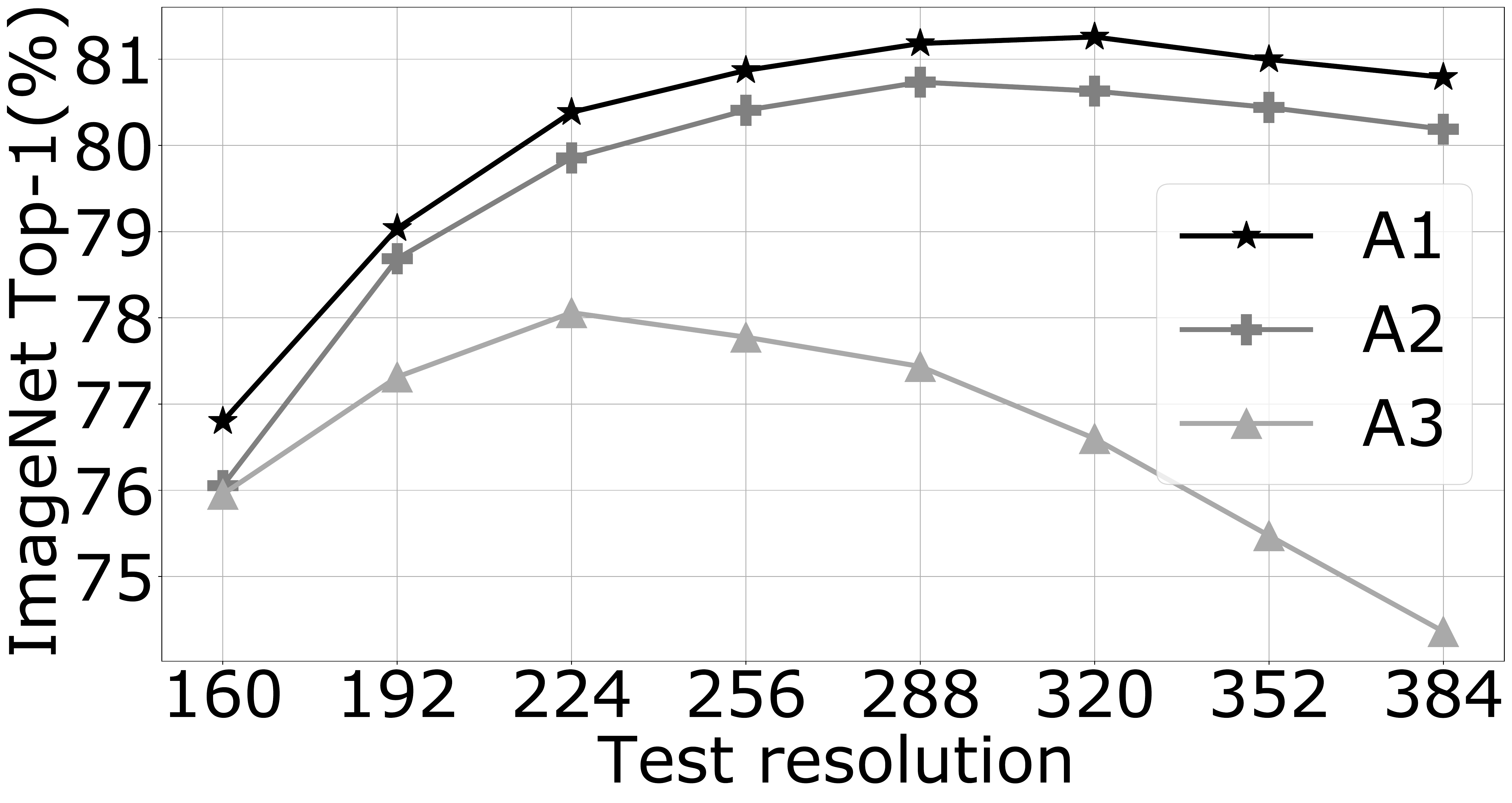}

    \caption{We compare ImageNet Top-1 accuracy according to the test resolution for our three training procedures A1, A2 and A3 with ResNet-50 architecture. %
    Our training procedure and models also benefit from the FixRes effect~\cite{Touvron2019FixRes}: the performance increases when using a larger image at test time for the procedures A1 and A2. This observation is not true for A3, which is expected since this procedure was already relying on feeding smaller images at train time, 
    so as to maximize the accuracy at test resolution $224\times224$. 
    \label{fig:opt_res}}
\end{figure}

While we primarily focus on the performance when inferring at resolution $224\times224$, we also evaluate our models when feeding images at larger resolutions. We report these results in Figure~\ref{fig:opt_res}, where we see that the models trained with A1 and A2 have a  better performance when used at higher resolutions. 

\section{Conclusion}
\label{sec:conclusion} 

In this paper we have proposed new training procedures for a vanilla ResNet-50. We have integrated new ingredients and put a significant effort in exploring diverse procedures under different resource constraints. As a result, we have established the new state of the art for training this gold-standard model. We have two other procedures to train strong ResNet-50 with less compute power. Nevertheless, we do not claim that our procedures are universal, quite the opposite:  the architecture and training should be optimized jointly. Our procedure is not ideal for training other models: while, on some models, our training recipes lead to excellent results outperforming those reported in the literature, they exhibit suboptimal performance on others, typically for deeper architectures that require more regularization.

\section*{Acknowledgments \& feedback}

Ross Wightman thanks NVIDIA for the donation of a V100 DGX Station and Google's TPU Research Cloud (TRC) for Cloud TPUs used in this research. All authors thank Mike Rabbat and Jakob Verbeek for their feedback. 
We welcome feedback regarding these or other noteworthy procedures via the \textbf{timm} GitHub Discussions\footnote{\url{https://github.com/rwightman/pytorch-image-models/discussions}} . 

\FloatBarrier
\newpage

\begingroup
    \small
    \bibliographystyle{ieee_fullname}
    \bibliography{egbib}
\endgroup

\clearpage

\appendix\newpage

\onecolumn

\begin{center}
{\LARGE
\textbf{Supplementary material} \\[0.2cm]}

\vspace{0.5cm} %
\end{center}

This supplemental material provides complementary results referred in the main document, noticeably a presentation of the augmentation and regularization specificity in \textbf{timm} and some alternative training procedures.

 \FloatBarrier

\FloatBarrier

\section{Augmentations and Regularization in timm~\cite{pytorchmodels}}
\label{sec:augreg}

The \textbf{timm} library includes a variety of image augmentations, regularization techniques, optimizers, and learning rate schedulers that can be used to produce leading results on ImageNet classification and other 2D image tasks. Many \textbf{timm} training components have modifications and improvements from original implementations or papers describing them. One should be aware of these changes if using them. 

\paragraph{Data Augmentation} in \textbf{timm} includes implementations of RandAugment~\cite{Cubuk2019RandAugmentPA}, AutoAugment~\cite{Ekin2018AutoAugment}, AugMix~\cite{Hendrycks2020AugMixAS}, Random Erasing~\cite{Zhong2020RandomED}, and an integrated implementation of Mixup~\cite{Zhang2017Mixup} and CutMix~\cite{Yun2019CutMix}. The base for all augmentations is typically Random Resized Crop with horizontal flipping. 

\paragraph{RandAugment} is the most used of the AA (AutoAugment) variants in \textbf{timm} -- it also contains the most significant additions from the original paper and Tensorflow based implementations – so we will focus on that implementation. The original RandAugment specification has two hyper-parameters, M and N; where M is the distortion magnitude and N is the number of distortions uniformly sampled and applied per-image. The goal of RandAugment was that both M and N be human interpretable. However, that ended up not being the case for M. The scales of several augmentations were backwards or not monotonically increasing over the range such that increasing M does not increase the strength of all augmentations.

This is most visible for image enhancement blending operations (color, contrast, brightness, sharpness) where the argument value defines the behavior as follows: 

\begin{minipage}{0.8\linewidth}
\begin{itemize}
    \item [0.] selects the degenerate image
    \item [0.-1.0] interpolates between the degenerate and original image
    \item [1.0] returns original image
    \item [$>$ 1.0] extrapolates the original image way from the degenerate
\end{itemize}
\end{minipage}

Taking sharpness as an example, magnitudes of M0, M5, and M10 are mapped in the original implementation to produce strong blurring (0.1), no-change (1.0), or strong sharpening (1.9) respectively.

The implementation in \textbf{timm} attempts to improve this situation by adding an ‘increasing’ mode (always enabled for recipes in this paper) where all augmentation strengths increase with magnitude; solarize and posterize increase with M (instead of decrease), and interpolation vs extrapolation for the blending operations is randomly chosen with a strength that increases with M. This makes increasing M more intuitive and allows an additional hyper-parameter to work well: \textbf{timm} adds a MSTD parameter which adds gaussian noise with the specified standard deviation to the M value per distortion application. Additionally, if MSTD is set to ‘-inf’, M is uniformly sampled from 0-M for each distortion. Without correcting the scales, one would often end up with completely empty or heavily inverted images in ranges of M that are supposed to be low in strength.

Care was taken in \textbf{timm}’s RandAugment to reduce impact on image mean, the normalization parameters can be passed as a parameter such that all augmentations that may introduce border pixels can use the specified mean instead of defaulting to 0 or a hard-coded tuple as in other implementations. And lastly, Cutout is excluded by default to favour separate use of \textbf{timm}’s Random Erasing implementation which has less impact on mean and standard deviation of the augmented images. 

\paragraph{Random Erasing} is another commonly used \textbf{timm} augmentation with modifications from the original paper. The implementation in timm follows the original but allows ‘erasing’ image regions with per-pixel gaussian noise (mean 0, std 1.0) instead of a uniform random or constant color (black or image mean) per-region. When applied to images at the recommended location in the augmentation pipeline -- after images have been normalized (standardized) -- this maintains image statistics and allows better results with stronger application of the augmentation. A count parameter was also added to \textbf{timm}’s Random Erasing such that multiple regions can be erased per-image. 

\paragraph{Mixup and CutMix} are cleanly integrated in \textbf{timm} in a manner not common in other implementations. Both can be enabled at the same time with a variety of different mixing strategies: 

\begin{description}
    \item [batchwise] CutMix vs Mixup selection, lambda and CutMix region sampling performed per-batch 
    \item [pairwise] mixing, lambda, and region sampling performed per mixing sample pair within batch
    \item [elementwise] mixing, lambda, and region sampling performed per sample within batch
    \item [half] the same as elementwise but one of each mixing pair is discarded so that each 
    sample is seen once per epoch
\end{description}

The default is to use either CutMix or Mixup with probability of 0.5 per-batch if both are enabled – this is the case for all mentioned training procedures in this paper. 

\paragraph{Regularization} in \textbf{timm} is standard. It allows use of similar regularization for many of the included models. Weight decay is available via either native PyTorch or \textbf{timm} optimizers. The ability to enable pre-classifier dropout is included in all model architectures. Stochastic-Depth has been added as an option to many of the most popular model architectures (via a layer named DropPath). Label-smoothing is included via a cross-entropy loss function and possible to use in combination with the label manipulation of CutMix and Mixup.

\section{Alternative Training Procedures}
\label{sec:alternative}

The main training recipes in this paper uses the LAMB optimizer. Several sets of hyper-parameter variations with differing training costs were presented with leading results for the vanilla ResNet-50 architecture. Here, we introduce alternative training recipes that also produce results matching or exceeding the best existing ResNet-50. The reader may find these are better suited for use or adaptation for their specific model architecture, dataset, or task. The alternative recipes are: 
\begin{description}
    \item [Procedure \Bp~ --] RMSProp with EMA %
    weight averaging and step LR decay;
    \item [Procedure \Cp~ --] SGD with Nesterov’s momentum, Adaptive Gradient Clipping, and a cosine learning rate decay. We have two variants of it (C1 and C2) depending on whether we use repeated augmentation or not; 
    \item [Procedure \Dp~ --] AdamP with a cosine learning rate decay and binary cross-entropy. 
\end{description}

The above procedures have been used to product excellent results for many pre-trained models in the \textbf{timm} library, including many non-ResNet architectures. Table~\ref{tab:others_training} summarizes their best ResNet-50 oriented settings. 

\begin{table}[t]
\centering
\scalebox{0.85}
{%
\begin{tabular}{l|cccc}
\toprule
Procedure $\rightarrow$ & \Bp & \Cp.1 & \Cp.2 & \Dp \\
\midrule
Train Res & 224 & 224 & 224 & 224 \\
Test Res & 224 & 224 & 224 & 224 \\
\midrule
Epochs  & 600 & 800 & 800 & 600 \\
\# of forward pass & 375k & 500k & 500k & 2,000k \\
\midrule
 Batch Size & 2048 & 2048 & 2048 & 384 \\
 Optimizer & RMSProp & SGD & SGD & AdamP \\
 Initial LR & 0.18 & 0.88 & 0.88 & 0.0033 \\
 LR Scheduler & step & cosine & cosine & cosine \\
 Decay Rate & 0.988 per 1-epoch & \xmarkg & \xmarkg & \xmarkg \\
 LR Noise (\% of training) & 0.45 to 1.0 & \xmarkg & \xmarkg & \xmarkg \\
  Weight Decay & $7.0\times 10 ^{-6}$ & $1.0\times 10 ^{-5}$ & $1.0\times 10 ^{-5}$ & 0.01\\
  Warmup Epochs & 5 & 5 & 5 & 5 \\
  
  \midrule
  Label Smoothing & 0.1 & 0.1 & 0.1 & 0.1 \\
  Dropout & 0.2 & 0.25 & 0.25 & 0.1 \\
  Stochastic Depth & 0.1 & 0.1 & 0.1 & 0.05 \\
  Repeated Augmentation & \xmarkg & \xmarkg & \cmark & \xmarkg \\
  Grad Clipping & \xmarkg  & AGC .025 & AGC .05 & \xmarkg  \\
  
  \midrule
  RandAugment (M/N/MSTD) & 8/2/1.0 & 7/3/1.0 & 7/3/1.0 & 7/3/1.0 \\
  Mixup & 0.2 & 0.2 & 0.2 & 0.2 \\
  CutMix & \xmarkg & 1.0 & 1.0 & 1.0 \\
  Random Erasing (Prob/Count) & 0.35/3 & 0.4/1 & 0.4/1 & .35/1 \\ 
  \midrule
  EMA weight averaging & 0.9999 & \xmarkg & \xmarkg & \xmarkg \\ 
  \midrule
  CE loss & \cmark & \cmark & \cmark & \xmarkg \\ 
  BCE loss  & \xmarkg & \xmarkg & \xmarkg & \cmark \\ 
  \midrule
  Top-1 acc. & 79.4\% & 79.8\% & 80.0\% & 79.8\% \\

 \bottomrule
\end{tabular}}
\smallskip
\caption{Alternative training procedures giving good performance with ResNet-50 architecture. 
\label{tab:others_training}}
\end{table}

\paragraph{Training procedure \Bp~ details (RMSProp)}
This procedure is inspired by the RandAugment~\cite{Cubuk2019RandAugmentPA} recipes used to train EfficientNet architectures but leverages features in \textbf{timm}’s implementation of RandAugment and Random Erasing. The step decay has been adjusted to decay every epoch (instead of every 2.4 as with EfficientNet, weight decay has been slightly decreased from EfficientNet defaults, and the learning rate is a bit higher. Additional augmentation was added in the form of per-pixel noise Random Erasing and Mixup. It should be noted that the RMSProp optimizer used is the \textbf{rmsprop\_tf} implementation in \textbf{timm} which carefully matches behaviours of the Tensorflow (before version 2.0) implementation. The native PyTorch RMSProp implementation will not produce the same results, even if adjusting for the epsilon location.

With long decay constants for the EMA weight averaging, it can be beneficial to perturb the learning rate (currently once per epoch) with noise in later stages of training (typically 40-50\% of the way through until the end). In exploration so far, learning rate noise appears to increase sensitivity of training results to random seed but has often produced the best result in (so far, limited) sweeps with the same hyper-parameters. Further analyzing the interplay between learning rate value, schedule, and noise, EMA decay constant, and random seed is a future objective for refining this training recipe.

This training strategy varies somewhat in effectiveness with batch size. Running experiments for this paper with larger batch sizes in the 1024-2048 range has often come slightly below (0.1 to 0.3 top-1) prior training runs with smaller sizes in the 256-768 range used for numerous \textbf{timm} pre-trained weights. It is unclear if this can be addressed with further hyper-parameter adjustments and different learning rate scaling (linear used by default) across batch sizes.

See Table~\ref{tab:procedure_b} for a summary of the procedure, including ranges of recommended of values to search over for applying to different classification task and architecture combinations. For larger model architectures it is advisable to focus on stronger augmentation and regularization values within the suggested ranges. Looking at Table~\ref{tab:mainres}, the original results for the \textbf{timm} specific EfficientNetV2-S~\cite{Tan2021EfficientNetV2SM} variant and ECA-ResNet-269-D were trained using this procedure, but with higher levels of augmentation and regularization than for ResNet-50.

\begin{table}[p]
\centering
\scalebox{0.85}
{%
\begin{tabular}{l|ll}
\toprule
 &  Recommended Range & ResNet-50 \\
 \midrule
 Epochs & 400-700 & 600 \\
 
 \midrule
 Initial LR (per batch size 256) & .01-.025 & 0.0225 \\
 LR Schedule & Step & Step \\
 LR Decay Rate & 0.97-0.99 per 1-3 epochs & 0.988 per 1-epoch \\
 LR Noise Active (\% of training) & 40-50\% to 100\% & 45\% to 100\% \\
 \midrule
 Grad Clipping & Off, global norm 1.0 & off \\
 \midrule
  Dropout & 0-0.4 & 0.2 \\ 
 Stoch Depth & 0-0.1 & 0.1 \\
 Repeated Augmentation & Off & Off \\
 \midrule
 RandAugment (M / N / MSTD) & 6-9 / 2-4 / 0.5-1.0 & 8 / 2 / 1.0 \\
 Random Erasing (Prob / Count) & 0.1-0.5 / 1-3 & 0.35 / 1 \\ 
  Mixup & 0.2, 0.5, 0.8 & 0.2 \\ 
 CutMix & Off, 0.8, 1.0 & 0 \\
 \midrule

 EMA Weight Averaging & On & On \\
 \midrule
 Loss & CE & CE \\ 

 \bottomrule
\end{tabular}}
\smallskip
\caption{Procedure \Bp~ summary 
\label{tab:procedure_b}}
\end{table}

\paragraph{Training procedure \Cp ~details (SGD with Nesterov's momentum and AGC)}

This recipe is based on the published procedure for training NFNets~\cite{Brock2021HighPerformanceLI}: using SGD with Nesterov’s momentum, Adaptive Gradient Clipping ~\cite{Brock2021HighPerformanceLI} (AGC), and heavy augmentation and regularization. AGC allows for stable large batch training at higher learning rates. Stronger default augmentation and regularization make up for the loss of batch normalization’s regularizing effect when paired with NFNets, but strengths can be relaxed when used with other architectures that use batch normalization. With some adjustments, this procedure has been useful training architectures in \textbf{timm} such as ECA-NFNet, ResNet, and EfficientNet variants to impressive performance levels. The original result for the \textbf{timm} EfficientNetV2-M~\cite{Tan2021EfficientNetV2SM} variant in Table~\ref{tab:mainres} was trained with the C.1 recipe, but with significantly higher augmentation and regularization than for ResNet-50.

The C.1 vs C.2 versions of this procedure seen in Table~\ref{tab:others_training} differ most significantly in the application of Repeated Augmentation. It should be noted that a shorter training length of 600 epochs also works quite well in both cases, with an expected drop of roughly 0.15-0.2 top-1 for the same seed. Table~\ref{tab:procedure_c} includes ranges of the ingredients for exploring with different tasks and architectures.

\begin{table}[p]
\centering
\scalebox{0.85}{
\begin{tabular}{l|ll}
\toprule
 &  Recommended Range & ResNet-50 \\
 \midrule
 Epochs & 300-800 & 800 \\
 \midrule
 Initial LR (per batch size 256) & .08-.12 & 0.11 \\
 LR Schedule & cosine & cosine \\
 \midrule
 Grad Clipping & AGC .01 - .05 & AGC .05 \\ 
 \midrule
  Dropout & 0-0.5 & 0.25  \\ 
 Stoch. Depth & 0-0.2 & 0.1  \\
 Repeated Augmentation & Off, On & On \\
 \midrule
 RandAugment (M / N / MSTD) & 6-10 / 2-4 / 0.5-1.0 & 7 / 3 / 1.0 \\
 Random Erasing (Prob / Count) & 0.1-0.5 / 1-3 & 0.4 / 1 \\ 
  Mixup & 0.2, 0.5, 0.8 & 0.2 \\ 
 CutMix & Off, 0.8, 1.0 & 1.0  \\
 \midrule
 Loss & CE & CE \\ 
 \bottomrule
\end{tabular}
}
\smallskip
\caption{Procedure \Cp~ summary 
\label{tab:procedure_c}}
\end{table}

\paragraph{Training Procedure \Dp ~ details (AdamP)}

Late in the process for this report a training trial using AdamP~\cite{adamp} showed promise. With limited runs so far a recipe based on AdamP has achieved a 79.8 top-1 on ImageNet-1k. Further experimentation is necessary, the trials so far were run at a comparatively small batch size, but the promising results warrant exploration. Note that unlike RMSProp or SGD (but similar to Adam and LAMB), it is recommended to use square root scaling when adjusting the learning rate for this recipe across different batch sizes.

Table~\ref{tab:procedure_d} contains the recommended ranges for the ingredients of this recipe. These ranges have not been explored extensively across different model architectures as with procedures B and C.

\begin{table}[p]
\centering
\scalebox{0.85}{
\begin{tabular}{l|ll}
\toprule
 &  Recommended Range & ResNet-50 \\
 \midrule
 Epochs & 300-600 & 600 \\
 \midrule
 Initial LR (per batch size 256) & .002 - .003 & 0.0027 \\
 LR Schedule & cosine & cosine \\
 \midrule
 Grad Clipping & None & None \\ 
 \midrule
  Dropout & 0-0.3 & 0.1  \\ 
 Stoch. Depth & 0-0.1 & 0.05 \\
Repeated Augmentation & Off, On & Off \\
 \midrule
 RandAugment (M / N / MSTD) & 6-9 / 2-4 / 0.5-1.0 & 7 / 3 / 1.0 \\
 Random Erasing (Prob / Count) & 0.1-0.5 / 1-3 & 0.35 / 1\\ 
  Mixup & 0.2, 0.5, 0.8 & 0.2 \\ 
 CutMix & Off, 0.8, 1.0 & 1.0  \\
 \midrule
 Loss & CE, BCE & BCE  \\ 
 \bottomrule
\end{tabular}
}
\smallskip
\caption{Procedure \Dp~ summary 
\label{tab:procedure_d}}
\end{table}

\paragraph{Other Recipes}

Undoubtedly, other training recipes with different combinations of optimizer, learning rate schedule, augmentation, and regularization exist that can match or surpass the performance of the procedures detailed in this report. Ingredients aside, putting in the time and effort to tune the recipe with the target architecture is key. The authors already have an AdamW recipe in the works that is looking promising.

\end{document}